\documentclass[sigconf]{acmart}
\usepackage{booktabs} % For formal tables
\renewcommand\footnotetextcopyrightpermission[1]{} % removes footnote with conference information in first column
\usepackage{url}
\usepackage{multirow}
\usepackage{graphicx}
\usepackage{comment}
\usepackage{subfigure}
\usepackage{dblfloatfix}
\usepackage{hyperref}
\usepackage{cleveref}

\setcopyright{none}
\begin{document}
\title[Diversity-enabled parsimonious  ensembles]{Developing parsimonious ensembles using ensemble diversity within a reinforcement learning framework}
%\titlenote{Produces the permission block, and
%  copyright information}
%\subtitle{Extended Abstract}
%\subtitlenote{The full version of the author's guide is available as
%  \texttt{acmart.pdf} document}

\author{Ana Stanescu}
%\authornote{}
\affiliation{%
  \institution{Department of Computer Science\\University of West Georgia}
%  \streetaddress{P.O. Box 1212}
  \city{Carrolton}
  \state{GA}
  \postcode{30118}
}
\email{astanesc@westga.edu}

\author{Gaurav Pandey}
%\authornote{}
\affiliation{%
  \institution{Department of Genetics and Genomic Sciences\\Icahn Institute for Genomics and Multiscale Biology\\Icahn School of Medicine at Mount Sinai}
%  \streetaddress{P.O. Box 1212}
  \city{New York}
  \state{NY}
  \postcode{10029}
}
\email{gaurav.pandey@mssm.edu}

\renewcommand{\shortauthors}{Stanescu \& Pandey}

\begin{abstract}
Heterogeneous ensembles built from the predictions of a wide variety and large number of diverse base predictors represent a potent approach to building predictive models for problems where the ideal base/individual predictor may not be obvious. Ensemble selection is an especially promising approach here, not only for improving prediction performance, but also because of its ability to select a collectively predictive subset, often a relatively small one, of the base predictors. In this paper, we present a set of algorithms that explicitly incorporate ensemble diversity, a known factor influencing predictive performance of ensembles, into a reinforcement learning framework for ensemble selection. We rigorously tested these approaches on several challenging problems and associated data sets, yielding that several of them produced more accurate ensembles than those that don't explicitly consider diversity. More importantly, these diversity-incorporating ensembles were much smaller in size, i.e., more parsimonious, than the latter types of ensembles. This can eventually aid the interpretation or reverse engineering of predictive models assimilated into the resultant ensemble(s).
\end{abstract}

\begin{comment}
\begin{CCSXML}
<ccs2012>
<concept>
<concept_id>10010147.10010257.10010321.10010333</concept_id>
<concept_desc>Computing methodologies~Ensemble methods</concept_desc>
<concept_significance>500</concept_significance>
</concept>
<concept>
<concept_id>10002951.10003227.10003351</concept_id>
<concept_desc>Information systems~Data mining</concept_desc>
<concept_significance>300</concept_significance>
</concept>
</ccs2012>
\end{CCSXML}

\ccsdesc[500]{Computer systems organization~Embedded systems}
\ccsdesc[300]{Computer systems organization~Redundancy}
\ccsdesc{Computer systems organization~Robotics}
\ccsdesc[100]{Networks~Network reliability}
\end{comment}
\keywords{Predictive modeling, Ensemble learning, Ensemble diversity, Reinforcement learning}

\maketitle

\section{Introduction}

%Over the last few decades, large amounts and great variety of data have been generated to study complex biological processes and diseases. The number and variety of predictive methods able to address many such biomedical prediction problems have also grown substantially in recent years, thus promoting the ``crowdsourcing" paradigm. However, despite this growth, it is generally difficult to determine the best method to address several biomedical problems, partly due to incomplete knowledge of the target problem and also data issues like noise and missing data. A powerful approach to problems/situations like these is to integrate the complementary knowledge embedded in many diverse base predictors by constructing ensembles \cite{Rokach2009,Seni2010}, which have been very successful in producing accurate predictions for many biomedical prediction tasks \cite{Yang2010,Altmann2008,Khan2010,Pandey2010,Yu2013,Guan2008,Ward:2006}. 

Ensemble methods combining output of individual classifiers~\cite{Rokach2009,Seni2010} have been  successful in producing accurate predictions for many complex classification tasks~\cite{Shotton2011,Niculescu-Mizil2009,Yang2010,Altmann2008,Liu2012,Khan2010,Pandey2010}. The effectiveness of an ensemble is related to the [degree of] diversity among its constituent base predictors, as complete consensus (lack of diversity) cannot provide any advantage over any individual base predictor, and similarly, a lack of consensus (high diversity) is unlikely to produce confident predictions. Successful ensemble methods strike a balance between diversity and accuracy \cite{Kuncheva2003,Dietterich2000}. For example, popular methods like boosting \cite{Schapire2012} and random forest \cite{breiman2001random} generate this diversity by sampling from or assigning weights to training examples. However, they generally utilize a single type of base predictor, such as decision trees, to build the ensemble. Such homogeneous
ensembles may not be the best choice for difficult classification problems or data sets, where the ideal base prediction method is often unclear due to incomplete knowledge and/or data issues. 

A more potent approach in this scenario is to build ensembles from a wide variety and large number of {\it heterogeneous} base predictors, which naturally lends diversity to the resultant ensembles. Two commonly used methods for building such \emph{heterogeneous ensembles} are {\it stacking} \cite{Merz1999,Wolpert1992}, and {\it ensemble selection} \cite{Caruana2004,Caruana2006}. Recently, we showed that these methods are more effective than homogeneous ensembles and individual classification methods for complex prediction problems in genomics \cite{Whalen2013,Whalen2016,Stanescu:2017}. Other studies have produced similar results \cite{Altmann2008,Niculescu-Mizil2009}.

Ensemble selection is an especially promising approach, not only for improving prediction performance, but also because of its ability to select a collectively predictive subset, often a relatively small one, of all input base predictors. The parsimony of such an ensemble can be of even greater value for gaining domain knowledge, where it is not only important to learn an accurate (ensemble) predictor, but also interpret what novel insights can be provided about the target problem by the base predictors included in the ensemble. For instance, in the important biomedical problem of predicting protein function \cite{Pandey2006}, it is critical to identify the biological features or principles on the basis of which accurate predictions of protein function are made \cite{radivojac2013large}. It would be easier to reverse engineer a smaller (more parsimonious) ensemble than a much larger one, such as all the base predictors taken together, to identify such features or principles.

However, despite its potential utility, the most well-known algorithms proposed for ensemble selection \cite{Caruana2004,Caruana2006} are generally greedy and ad-hoc in nature. Specifically, due to the uncertainities in setting their parameters and components, the behavior and good performance of these algorithms is difficult to guarantee for a variety of problems or datasets. To address such challenges with these algorithms, we proposed a novel ensemble selection framework based on reinforcement learning (RL) \cite{Sutton:1998} in our previous work \cite{Stanescu:2017}. This framework provides a systematic way of exhaustively exploring the many possible combinations of base predictors that can be selected into an ensemble, and mathematically guarantees convergence to (one of) the optimal solutions. In our initial experiments \cite{Stanescu:2017}, the RL-based ensembles turned out to be almost as predictive as the much larger ensembles consisting of all the base predictors, while being substantially smaller, i.e., more parsimonious. 

The effectiveness of this framework can be partly attributed to its utilization of the diversity of base predictors owing to the RL-enabled systematic search of the base predictor space, even without diversity among predictors being explicitly used as a criterion for this search. 
Given the close relationship between ensemble performance and diversity discussed above, we investigated in the current work whether explicity considering  the diversity among base/ensemble predictors within our RL-based framework can help build even more accurate and parsimonious ensemble predictors. In Section 2, we describe the algorithms implementing this extension using established measures for ensemble diversity 
\cite{Kuncheva2003, Tang2006}. Section 3 describes the data sets and experimental setup used to evaluate these algorithms. In Section 4, we illustrate how, and under what circumstances, these algorithms can achieve their goals of building accurate and parsimonious ensemble predictors. We conclude with directions for future work in Section 5.

\section{Proposed Approach}
Here, we describe our proposed ensemble selection approach, starting with basics of reinforcement learning and how it can be utilized to address the ensemble selection problem.

\subsection{Basics of Reinforcement Learning (RL)}
\label{sectionRL}
RL is a type of machine learning in which an agent  learns how to optimally behave in its environment using a trial-and-error mechanism \cite{Sutton:1998}. The agent tries to maximize the cumulative reward that it is receiving from making decisions. Reinforcement signals are provided by the reward function, which gradually guides the agent towards finding the optimal behavior, or ``policy", that will maximize its performance. 
%{\bf Recommend cutting all this out: It means that the agent will receive corrective feedback in the form of a reward for taking an action \emph{a} from a state \emph{s}. The problem is formulated as a Markov Decision Process, a framework which states the current state contains sufficient information of the future, such that the advancement of the MDP is independent of the past given the current state. The reward function, as well as the finite sets of actions and states are deterministic, meaning that every time the agent chooses to take action \emph{a} from state \emph{s} it will transition to state \emph{a'} and it will receive the same exact reward. The agent's task is thus learning an optimal behavior, or ``policy" for obtaining the maximum reward, or more specifically, it's learning to estimate delayed cumulative reward when following some learned policy,}

\emph{Q-learning} is one RL algorithm that can discover an optimal policy in a theoretically sound way \cite{Watkins:1992}. Here, the Q-table represents the action-value mapping that the agent constructs by training in the environment. Q-learning  allows the agent to investigate the environment and update the Q-table using an $\epsilon$-greedy approach: with probability $1-\epsilon$, the agent exploits its accumulated knowledge (best current policy as determined by the Q-table), and with probability $\epsilon$ the agent will explore another state in the environment, different from the one dictated by its current knowledge. Typically, the exploration is purely random, meaning that the agent is allowed to randomly pick the next action to take as long as it is different from the 
``optimal'' action. This allows the agent to potentially discover better actions than the ones it has already assessed.

\begin{figure}
\includegraphics[height=2.5in]{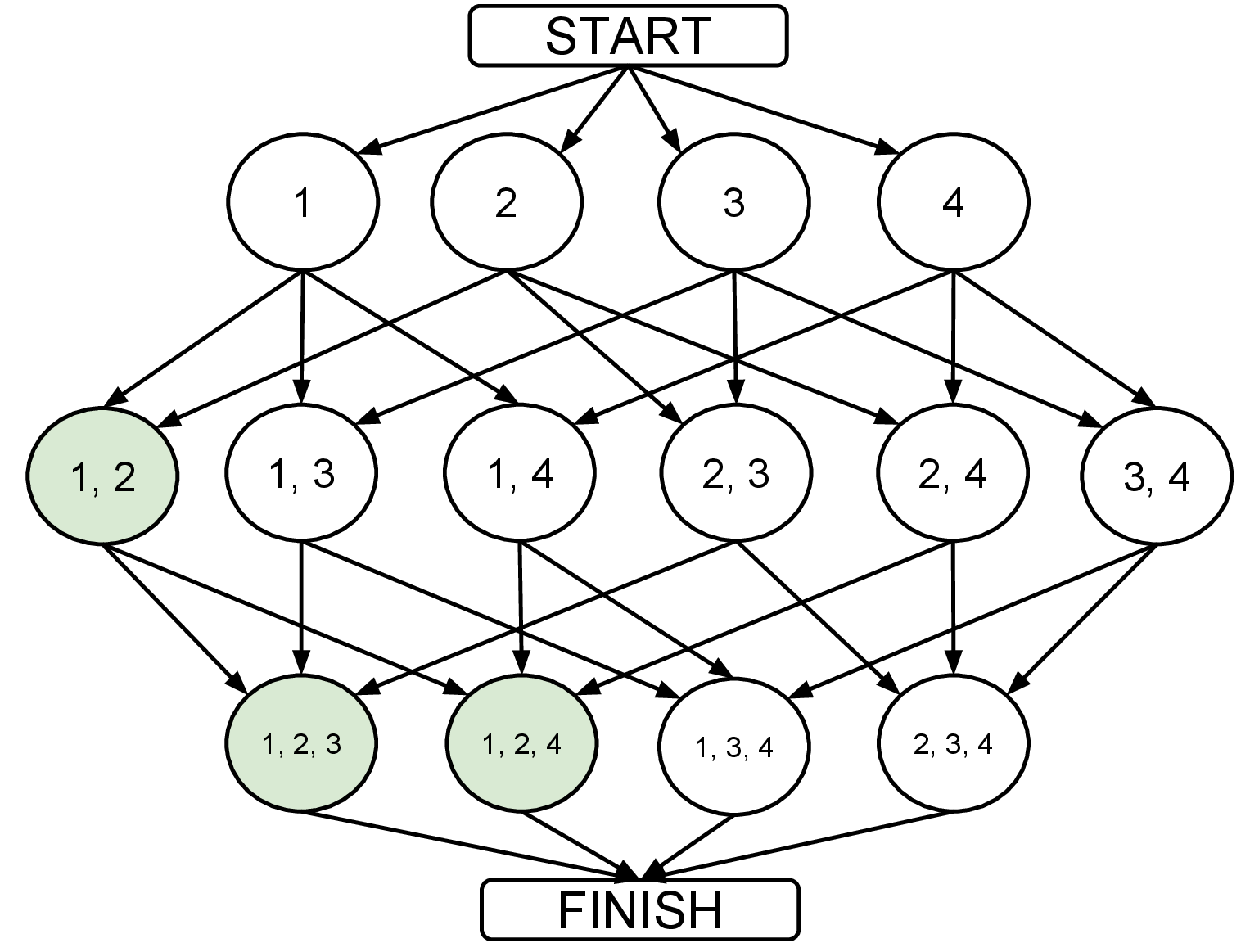}
\caption{An environment generated from 4 base predictors to be used for RL-based ensemble selection.}
\label{4-node_lattice}
\end{figure}

\subsection{Our previous work on RL-based ensemble learning}
In our previous work \cite{Stanescu:2017}, we formulated the ensemble selection problem as a search of the ensemble space and used Q-learning to train an agent, i.e., the ensemble selection algorithm. Specifically, we defined the RL agent's environment as a lattice, such as the one shown in Figure \ref{4-node_lattice}, where the nodes representing the states  denote all possible ensembles generated by N base predictors (\emph{i.e.}, ${2^N}$ possible nodes), and the arrows represent the allowed transitions between states (the ``actions'' in our setup).
We also proposed three strategies, RL\_greedy, RL\_backtrack and RL\_pessimistic, to search through this environment by defining the possible actions between states. This is done within the Q-learning framework to identify the best possible ensemble given the search strategy, the corresponding reward function and the value of $\epsilon$. Specifically, the framework returns the optimal policy, which in this case represents the optimal path from the root of the lattice to the selected ensemble.

%If an agent is located at a node determined by the ensemble \emph{(1, 2)} consisting of base predictors \emph{1} and \emph{2}, it can take N-2 possible actions and reach nodes determined by the ensembles consisting of base predictors \emph{(1, 2, 3), (1, 2, 4), ..., (1, 2, N-1)}, and \emph{(1, 2, N)}. For example, in Figure \ref{4-node_lattice}, the world is determined by all possible ensembles created with 4 base predictors. When located at node \emph{(1, 2)}, the agent can only take two actions, which can lead to states \emph{(1, 2, 3)} and \emph{(1, 2, 4)}, respectively.

In particular, the RL\_greedy strategy emulates a ``greedy'' agent, whose goal is to reach the state in the environment representing the full ensemble (FINISH node in Figure \ref{4-node_lattice}) and thus terminate each of its learning episodes as
quickly as 
possible. As a consequence of this plan, the agent rapidly takes actions to add the 
next base predictor to the ensemble
represented by the current 
state, and updates the Q-table in the process. With each action $a_{t}$ taken to transition from state $s_t$ to $s_{t+1}$ in the environment, the agent receives the performance of the ensemble represented by state $s_{t+1}$
as a reward (denoted by the function $f(s)$  in Equation \ref{equationS1reward}). 
\begin{equation}
{R(s_{t}, a_{t}, s_{t+1}) = f(s_{t+1})}
\label{equationS1reward}
\end{equation}

Finally, with all the episodes completed, and the optimal policy returned, RL\_greedy picks the final ensemble as the one on the policy path that produces the highest individual performance (on a validation set).

\begin{figure*}
\includegraphics[scale=0.45]{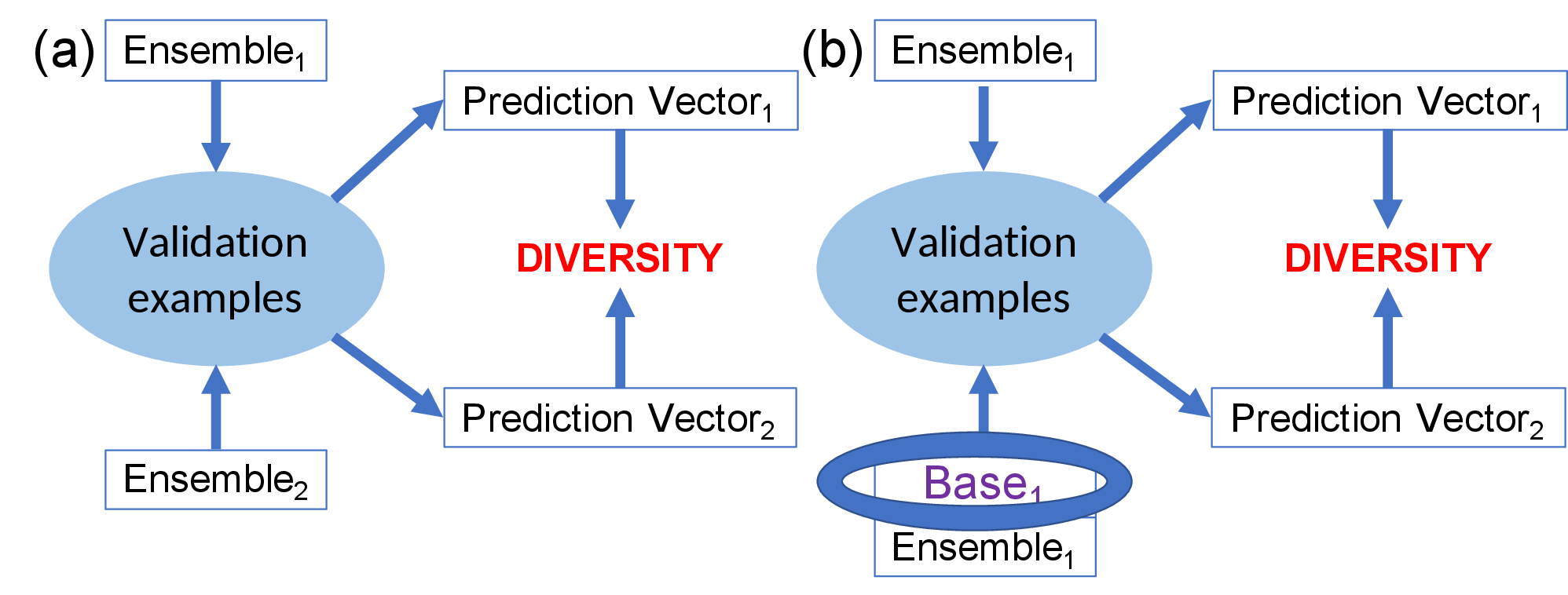}
	\caption{Diversity calculation methods used to quantify the diversity between (a) any pair of ensembles (\emph{diversity1}),
	and (b) a pair of ensembles differing in just one base predictor (\emph{diversity2}).}
\label{diversity_calculation_strategies}
\end{figure*}

\subsection{Incorporating ensemble diversity into our RL framework}
Our previously proposed RL strategies focus on systematically exploring the ensemble space that may help discover diverse and accurate ensembles. However, they don't explicitly consider diversity among predictors, as shown by the fact that their reward functions (e.g., Equation \ref{equationS1reward}), are only based on ensemble performance. Thus, given the close relationship between ensemble performance and diversity, in this work, we investigated if
the explicit incorporation of
ensemble diversity, 
measured appropriately, can help the RL framework explore and yield even more accurate and parsimonious ensembles.

The RL\_greedy strategy offers a natural extension towards a diversity-enabled version of the RL framework. Note that the natural form of Q-learning enables exploration of randomly chosen states with probability $\epsilon$. Thus, instead of random exploration, we can force the exploration to explicitly visit the state that is the \emph{most diverse} with respect to the ensemble represented by the current state. We adopted this approach 
and explain its exact implementations evaluated in our work below. The rest of the RL\_greedy strategy is used as it is.

\subsubsection{Diversity calculation methods}
Ensemble diversity has long been studied as a factor influencing ensemble performance, and several diversity measures have been proposed for this purpose \cite{Kuncheva2003, Tang2006}. However, most of these measures are defined for a pair of classifiers/predictors, which doesn't lend them naturally to calculating diversity involving at least one ensemble consisting of multiple base predictors. Specifically, in our case, the diversity needs to be calculated between candidate ensembles at the two ends of various transitions/actions in the RL environment, such as those denoted by green nodes and connected by the solid arrows in Figure \ref{4-node_lattice}. Thus, we used the following two simple methods for this calculation:
\begin{itemize}
\item \emph{diversity1}: The first method simply calculates the diversity measure between the output prediction vectors obtained by applying both the ensembles separately to the validation examples (Figure \ref{diversity_calculation_strategies}(a)).
\item \emph{diversity2}: The second method is based on the observation that the two ensembles only differ in one base predictor, such as predictor $3$ or $4$ with respect to the ensemble $\{1,2\}$ among the shaded nodes in Figure \ref{4-node_lattice}. Thus, in this strategy, only this distinct base predictor is applied to the validation examples to generate the prediction vector. The diversity measure is then calculated between this vector and that of the predictions generated by the original ensemble applied to the same examples (Figure \ref{diversity_calculation_strategies}(b)). 
\end{itemize}

\subsubsection{Ensemble diversity measures}
As mentioned before, a variety of diversity measures have been proposed for studying ensembles \cite{Kuncheva2003, Tang2006}. To test the effectiveness of incorporating diversity into our RL framework, we chose some representative measures of two types, \emph{unsupervised} and \emph{supervised}.\\

\noindent {\bf Unsupervised measures}: The simplest measures for the diversity of two sets of predictions are standard proximity measures, which are also unsupervised 
as they don't involve using the actual labels of the examples the predictions are generated from. In this work, we tested
Pearson's correlation coefficient, cosine similarity and Euclidean distance  for this purpose. Also, since diversity is analogous to distance, one minus the values of the correlation coefficient and cosine similarity were used as diversity measures.\\

\noindent {\bf Supervised measures}: The more established ensemble diversity measures are supervised in nature, as they don't only measure how similar or different a pair of set of predictions are, but also use the true label information of the examples to measure how similar or different their classification errors are. We tested Yule's $Q$  \cite{Yule1900} and Fleiss' $\kappa$ \cite{Dietterich2000} statistics as representatives of these measures in our study.
Specifically, 
given the predicted labels produced by a pair of predictors 
$D_i$ and 
$D_k$, a contingency table counting how often each predictor produces the correct label in relation to the other can be generated as:

\begin{table*}[t!]
\centering
\begin{tabular}{c|ccc|ccccc}
\multirow{3}{*}{Problem} & \multicolumn{3}{c|}{\begin{tabular}[c]{@{}c@{}}Protein Function Datasets (PF)\end{tabular}} & \multicolumn{5}{c}{\multirow{2}{*}{Splice Site Datasets (SS)}} \\
 & PF1 & ~~~~PF2 & PF3 & \multicolumn{5}{c}{} \\
 & \multicolumn{3}{c|}{({\it S. cerevisiae})} & {\it D. melanogaster} & {\it C. elegans} & {\it P. pacificus} & {\it C. remanei} & {\it A. thaliana} \\ \hline
\#Features & 300 & ~~~~300 & 300 & 141 & 141 & 141 & 141 & 141 \\
\#Positives & 382 & ~~~~344 & 327 & 1,598 & 997 & 1,596 & 1,600 & 1,600 \\
\#Negatives & 3,597 & ~~~~3,635 & 3,652 & 158,150 & 99,003 & 156,326 & 157,542 & 158,377 \\
\#Total & 3,979 & ~~~~3,979 & 3,979 & 159,748 & 100,000 & 157,922 & 159,142 & 159,977
\end{tabular}
\caption{Details of protein function (PF) and splice site (SS) datasets, including the number of features, number of examples in the minority (positive) and majority (negative) classes, and total number of examples.}
\label{tableDatasets}
\end{table*}

\begin{center}
\begin{tabular}{l|cc}
& $D_k$ correct (1) & $D_k$ incorrect (0) \\
\midrule
$D_i$ correct (1) & $N^{11}$ & $N^{10}$ \\
$D_i$ wrong (0) & $N^{01}$ & $N^{00}$ \\
\end{tabular}
\end{center}

Then, the $Q$ and $\kappa$ statistics can be calculated as:
\begin{align}
Q_{i,k} = \frac{N^{11} N^{00} - N^{01} N^{10}}{N^{11} N^{00} + N^{01} N^{10}} ~.
\end{align}
\begin{align}
	\kappa_{i,k} = \frac{2(N^{11} N^{00} - N^{01} N^{10})}{(N^{11}+N^{10})(N^{01}+N^{00})(N^{11}+N^{01})(N^{10}+N^{00})} ~.
\end{align}

These measures take the same scale as the correlation coefficient, i.e., they produce values tending towards~$1$ when~$D_i$ and~$D_k$ correctly classify the same instances,~$0$ when they do not, and~$-1$ when they are negatively correlated. So, we used one minus their respective values to quantify diversity.

Finally, like RL\_greedy that the diversity-incorporating algorithms are based on, the stopping point is reached when they produce the same ensemble for ten consecutive learning episodes.

\section{Evaluation Methodogy}
\label{sectionExperimentalSetup}
We evaluated our proposed diversity-enabled RL-based ensemble selection approaches, as well as their benchmarks, on two challenging computational genomics problems and several data sets related to them using the experimental setup described below.

\subsection{Problem Definitions and Datasets}
\noindent {\bf Protein Function Prediction}: 
Prediction of gene/protein function is an important problem in biology \cite{Pandey2006,radivojac2013large}. Gene expression data are commonly used for this task, as the simultaneous measurement of gene expression across the entire genome enables effective inference of functional relationships and annotations. Thus, we used the gene expression compendium of Hughes {\it et al.}\cite{Hughes2000} to predict the functions of roughly $4,000$ {\it S. cerevisiae} genes. Among these genes, the three most abundant functional labels (GO terms) from the list of most biologically interesting and actionable Gene Ontology Biological Process terms compiled by Myers {\it et al.} \cite{Myers2006} are used in our evaluation. These labels are GO:0051252 (regulation of RNA metabolic process), GO:0006366 (transcription from RNA polymerase II promoter) and GO:0016192 (vesicle-mediated transport). We refer to these prediction problems as PF1, PF2, and PF3 respectively (details in Table~\ref{tableDatasets}). 

\noindent {\bf Prediction of Splice Sites:}
RNA splicing is a naturally occurring biological phenomenon that contributes to protein diversity, which in turn contributes to the (ab)normal functioning of eukaryotic organisms \cite{black2003}.
Generally, when creating mature RNA from DNA, introns are removed (or spliced out) from the gene sequence and exons are retained (or transcribed). Splice sites are conserved nucleotide dimers found at the interfaces between exons and introns. In general, splice sites are {\it canonical}, as acceptor splice sites are signaled by the occurrence of the consensus dimer ``AG" at the $3'$ end of the intron, while donor splice sites are characterized by the consensus dimer ``GT", situated at the $5'$ end of the intron. Such dimers occur frequently throughout most eukaryotic genomes but their presence alone is not sufficient to declare a splice site. Correctly identifying splice sites is an essential step towards genome annotation, and a difficult problem due to the highly unbalanced ratio of bona fide splice sites to decoy dimers \cite{Shapiro:1987}. In this work, we focused on identifying acceptor splice sites, which can be formulated as a binary classification of DNA sequences (141-nucleotide-long windows around ``AG" dimers, with the dimer situated at position 61) as true acceptor splice sites and decoy dimers. Thus, we assessed the ability of our algorithms to address this important problem on five datasets of acceptor splice sites from five organisms: {\it D. melanogaster, C. elegans, P. pacificus, C. remanei}, and {\it A. thaliana}, published by Schweikert {\it et al.}\cite{Schweikert:2009} and R{\"a}tsch {\it et al.}\cite{Ratsch:2007}. 

% http://www.ncbi.nlm.nih.gov/pmc/articles/PMC1808025/
% https://www.semanticscholar.org/paper/Domain-Adaptation-in-Sequence-Analysis-Domain-Widmer-Huson/7461da7d0b606994985dc606d60325b9821eedbe/pdf
% DATA: http://cbio.mskcc.org/public/raetschlab/user/cwidmer/

\subsection{Experimental Setup}
We maintain the same experimental setup as used 
in our previous work \cite{Stanescu:2017}, illustrated by Figure 2 in that paper. Also, as in that work,
in all our RL-based experiments, the parameters of the Q-learning algorithm, namely  $\alpha$ and $\gamma$, are set to the commonly used values of 0.1 and 0.9 respectively \cite{Sutton:1998}. The exploration/exploitation trade-off is controlled by the $\epsilon$ probability discussed in Section \ref{sectionRL}. A higher probability indicates more exploration. In our work, we experiment with $\epsilon \in \{0.01, 0.1, 0.25, 0.5\}$. The iterative nature of Q-learning requires the initialization of its parameters (values in the Q-table). We initialize the Q-table as a $zero$ matrix, and update the values as states and rewards are observed by the agent, as guided by the ensemble diversity-based search strategies defined above.

We use 5-fold cross validation (CV) to estimate the performance of all the models. All base predictors are learned on the training set (60\% of the original data described in Table \ref{tableDatasets}), which is balanced using undersampling of the majority class to address the highly skewed distribution of the classes in our evaluation data sets. The validation set (20\% of the data) is used for calculating the rewards of the nodes in the RL environment 
as well as calculating the diversity measures described earlier. The test set (comprising the remaining 20\% of the data) is used to assess the overall performance of all studied algorithms. An experiment for an algorithm being tested consists of the collection of these performance scores over all five rounds of this cross-validation. 

\begin{figure*}[t!]
\centering
\begin{minipage}{.5\textwidth}\centering
  \centering
  %[trim = left bottom right top]
  \includegraphics[trim = 30mm 8mm 50mm 15mm, clip, width=\linewidth]{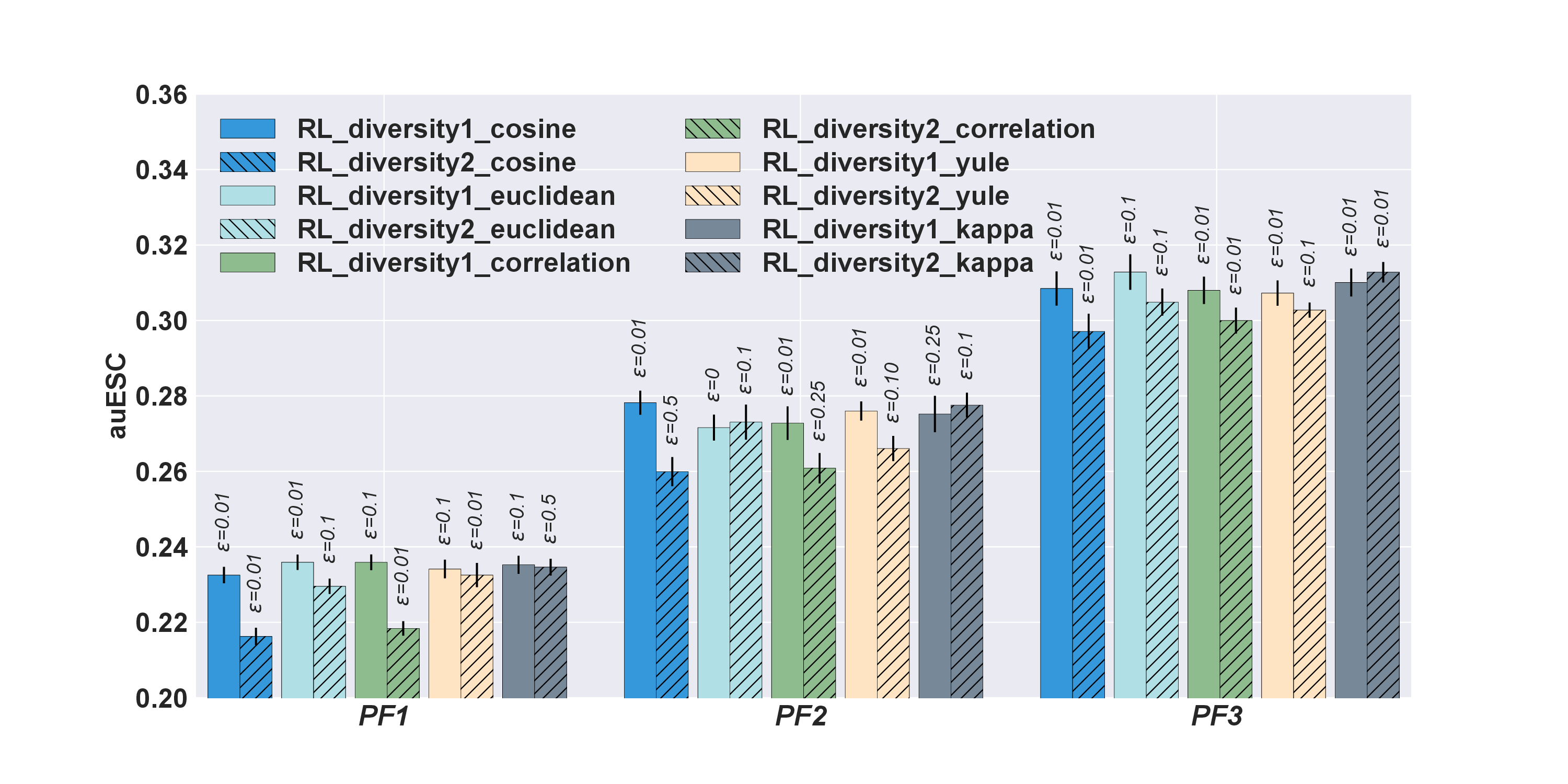}
  \normalsize{(a) Protein Function (PF) datasets}
\end{minipage}%
\begin{minipage}{.5\textwidth}\centering
  \centering
  \includegraphics[trim = 30mm 12mm 50mm 20mm, clip, width=1.0\linewidth]{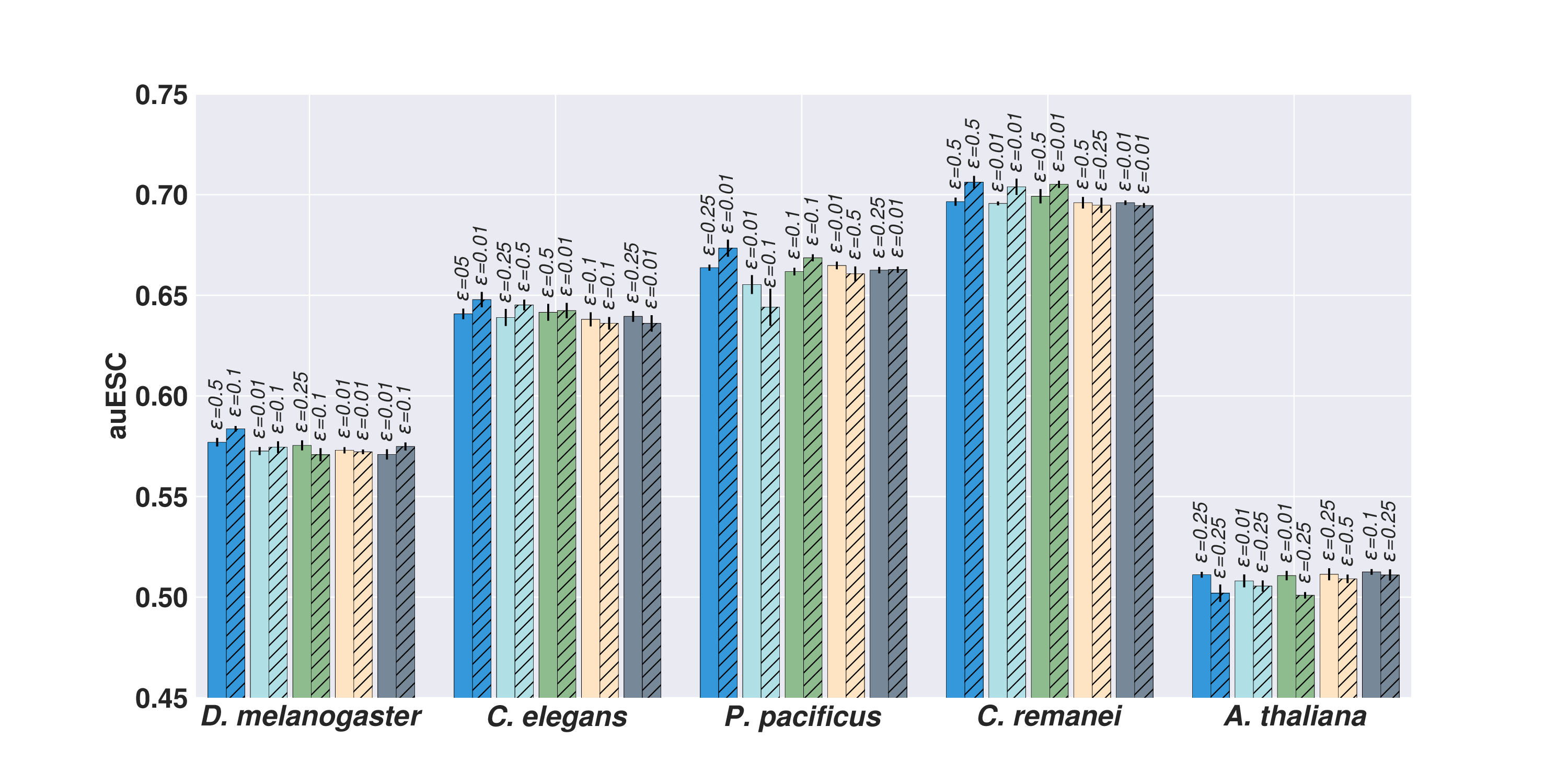}
  \normalsize{(b) Splice Site (SS) datasets}
\end{minipage}
\caption{Overall performance of all diversity-incorporating RL-based ensemble selection algorithms, computed as the auESC score described in Section \ref{sectionExperimentalSetup}, for (a) PF and (b) SS datasets. The same legend applies to both the plots. Standard errors are calculated over ten repetitions of all respective experiments. The $\epsilon$ value that produced the highest auESC score for each algorithm is shown above the corresponding bar.}
\label{diversity_barplots}
\end{figure*}

All performance evaluation, whether internal (on the validation set) or external (on the test set) is conducted using F-max, which is the maximum value of the F-measure across all the values of precision and recall at many thresholds applied to the prediction scores generated by the base classifiers and the resultant ensembles. F-max is appropriate given the highly skewed class distributions of the datasets used in our study, and has been shown to be reliable for  performance evaluation in a recent large-scale assessment of protein function prediction \cite{radivojac2013large}.

The ensembles selected by the various algorithms we tested are created by combining the probabilities produced by the constituent base predictors using a weighted average. Here, the importance (weight) of each base predictor is proportional to its predictive performance (measured in terms of the F-max score) on the validation set for all the approaches. We also considered options such as unweighted mean and median, but observed that the performance of the ensembles was suffering because of the worst performing individual base predictors included in them.
In order to efficiently obtain the reward of each state or the performance of the ensemble being considered, we aggregate the predictions using a cumulative moving average. 

We train 18 diverse representative base classification algorithms from Weka \cite{Weka}, including Na{\"i}ve Bayes, Multilayer Perceptron, SVM with a polynomial kernel, AdaBoost, Logistic Regression, and Random Forest. Each CV training set is resampled with replacement 10 times to balance the classes, resulting in 180 base predictors. Each ensemble selection algorithm is presented with a pool of base predictors (classification models). We start all our experiments with ten base predictors and increase this set gradually, with steps of ten randomly selected base predictors for each experiment, until we reach the entire set of 180 base predictors. This setup is designed to address the question of how the ensemble selection methods behave with an increasingly larger set of initial base predictors to select from. The performance of the ensembles resulting from of all these methods was evaluated across all these sizes, resulting in curves such as the ones shown in Figure \ref{RL_comparison_ramenei}. We used the area under these curves, denoted \emph{auESC} (\emph{area under Ensemble Selection Curve}), as a global evaluation measure.  To account for variation, each set of experiments was repeated ten times for all the datasets. Note that, in contrast to the ordering strategy used in our previous work \cite{Stanescu:2017}, the ordering of base predictors to generate increasingly larger sets to present to the ensemble selection algorithms is now completely random, thus reducing the likelihood of biasing the results.

\section{Results}
In this section, we describe the results of the evaluation of the algorithms described in Section 2 using the experimental setup described in Section 3. Note that in some of the discussions below, we discuss the PF and SS results separately due to the substantially different sizes of these datasets to study the effect of this important factor on the results.

\subsection{Performance of proposed algorithms}
As explained in Section 2, our diversity-incorporating RL-based ensemble selection algorithms have several variations, particularly because of the possible options for the following two components:
\begin{itemize}
\item Diversity computation methods: \emph{diversity1} or \emph{diversity2}.
\item Ensemble diversity measures: Unsupervised (\emph{cosine} similarity, \emph{Euclidean} distance and Pearson's \emph{correlation} coefficient) and supervised (\emph{Yule}'s Q and $\kappa$ (\emph{kappa})).
\end{itemize}
Thus, we first evaluated the performance of these variations using the data sets and experimental setup described in Section 3. The results of this evaluation are shown in Figure \ref{diversity_barplots}. Note that for all the variations, the value of $\epsilon$, the parameter defining the RL exploration-exploitation probability, that produces the highest auESC score for each data set is also shown. This enables a comparison between the best performing versions of the tested algorithms.

Overall, it can be seen that the general performance on the PF data sets (Figure \ref{diversity_barplots}(a)) is substantially lower than that on the SS ones (Figure \ref{diversity_barplots}(b)). This can be attributed to the larger size of the latter, which provides the ensemble algorithms more information to leverage for better predictions. 

\begin{figure*}[t!]
\centering
\subfigure[\normalsize{Performance on PF data sets}]{\includegraphics[trim = 30mm 12mm 50mm 20mm, clip, scale=0.25]{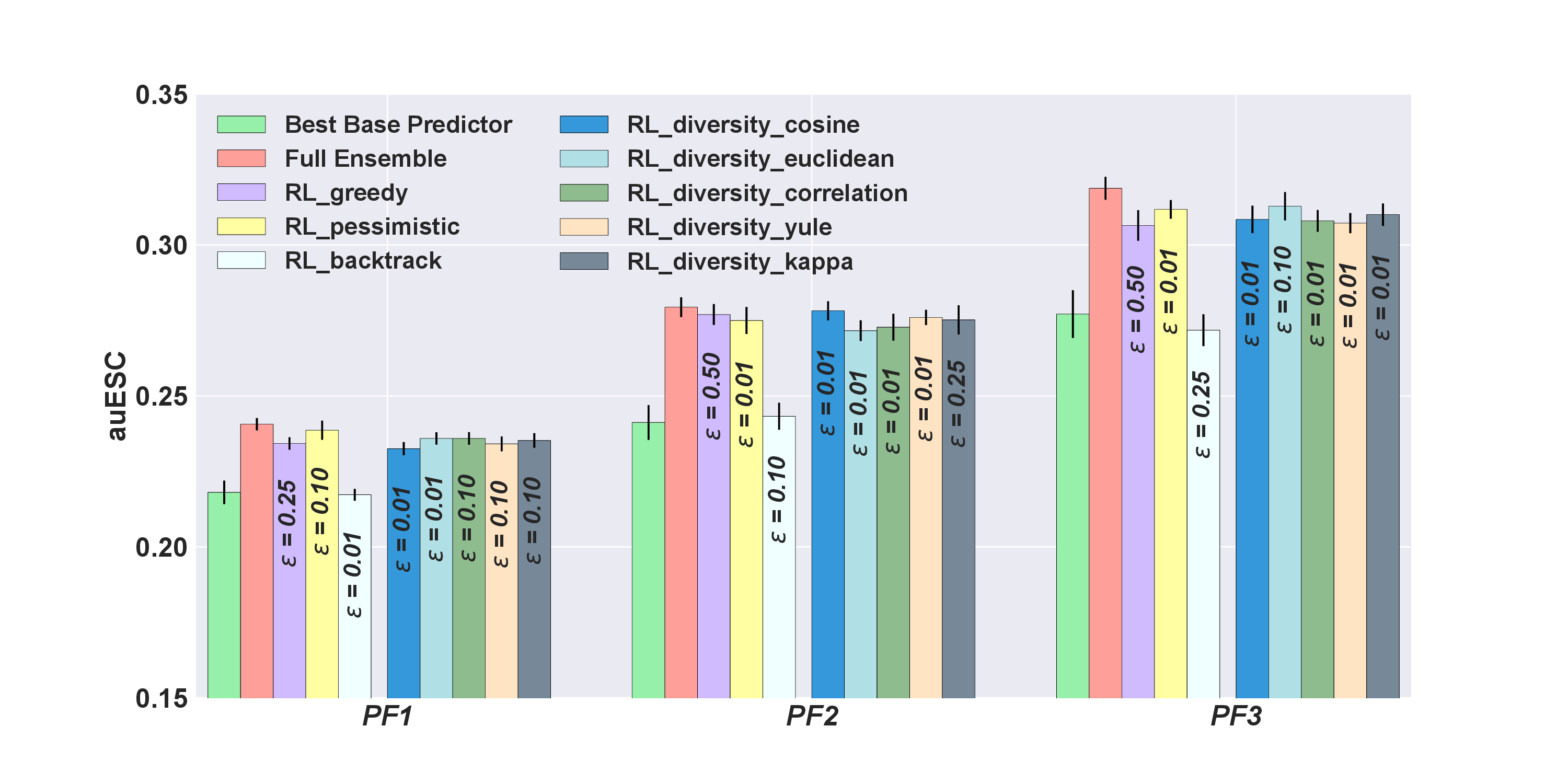}}
\subfigure[\normalsize{Performance on SS data sets}]{\includegraphics[trim = 30mm 12mm 50mm 20mm, clip, scale=0.32]{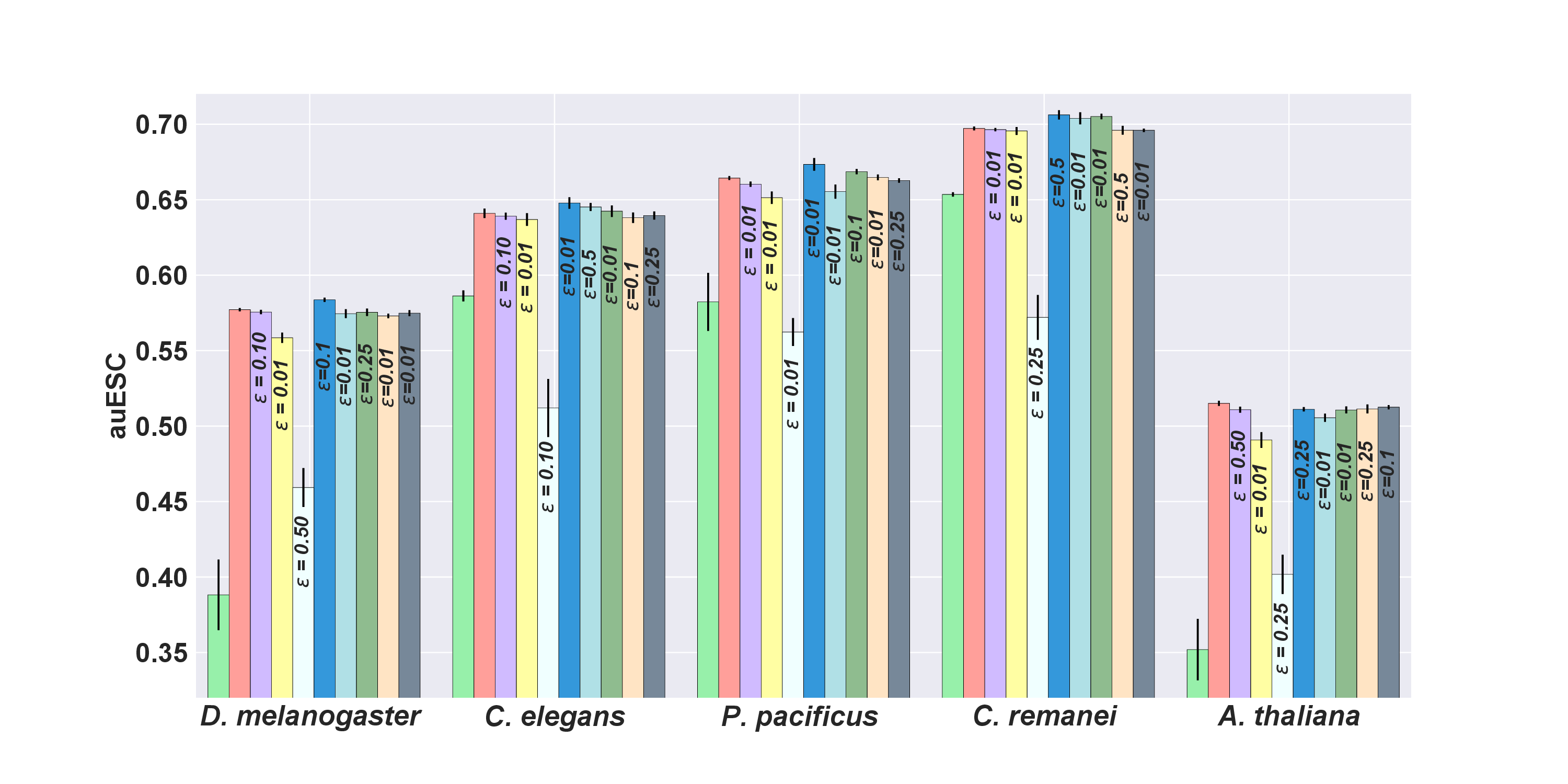}}
\caption{Performance of the diversity-incorporating RL-based ensemble selection algorithms, as well as those proposed in our previous work \cite{Stanescu:2017}, on (a) PF and (b) SS data sets. The same legend applies to both the plots. Also shown for reference are the performance of the full ensemble comprising of all the base predictors, as well as the best one among them. Standard errors are calculated over ten repetitions of all the respective experiments. The \emph{diversity1} method was used to generate the PF results, while the best performing one out of \emph{diversity1} and \emph{diversity2} was use to generate the SS ones for each data set and diversity measure. The $\epsilon$ value that produced the highest auESC score for each algorithm is shown within the corresponding bar.} 
\label{RL_comparison}
\end{figure*}

Next, we analyzed which of the \emph{diversity1} or \emph{diversity2} methods, whose performance is represented by paired solid and striped bars respectively in Figure \ref{diversity_barplots}, is more effective for diversity incorporation. It's quite clear from Figure \ref{diversity_barplots}(a) the \emph{diversity1} almost consistently performs better \emph{diversity2} for PF data sets (paired Wilcoxon rank-sum p-value=$0.0015$). However, in the case of the SS data sets (\ref{diversity_barplots}(b)), the performance of the two methods is statistically tied (paired Wilcoxon signed-rank p-value=$0.19$). Thus, for further analyses, for each diversity measure, whichever of the two methods produces the highest auESC score for each SS data set is used as the representative performance for that diversity measure for that dataset.

Finally, we analyzed if using unsupervised and supervised ensemble diversity measures had different effects on the performance of the above ensemble selection algorithms. For this, we statistically compared the average performance of these two groups of measures coupled with the \emph{diversity1} and \emph{diversity2} methods over all the PF and SS data sets using the paired Wilcoxon signed-rank test. However, we found there to be no significant differences ($p-value>0.1$ for all comparisons) between the two groups for any dataset type or diversity incorporation method. 

\begin{figure*}[!b]
\centering
%trim = 30mm 12mm 50mm 20mm, clip, 
\includegraphics[scale=1]{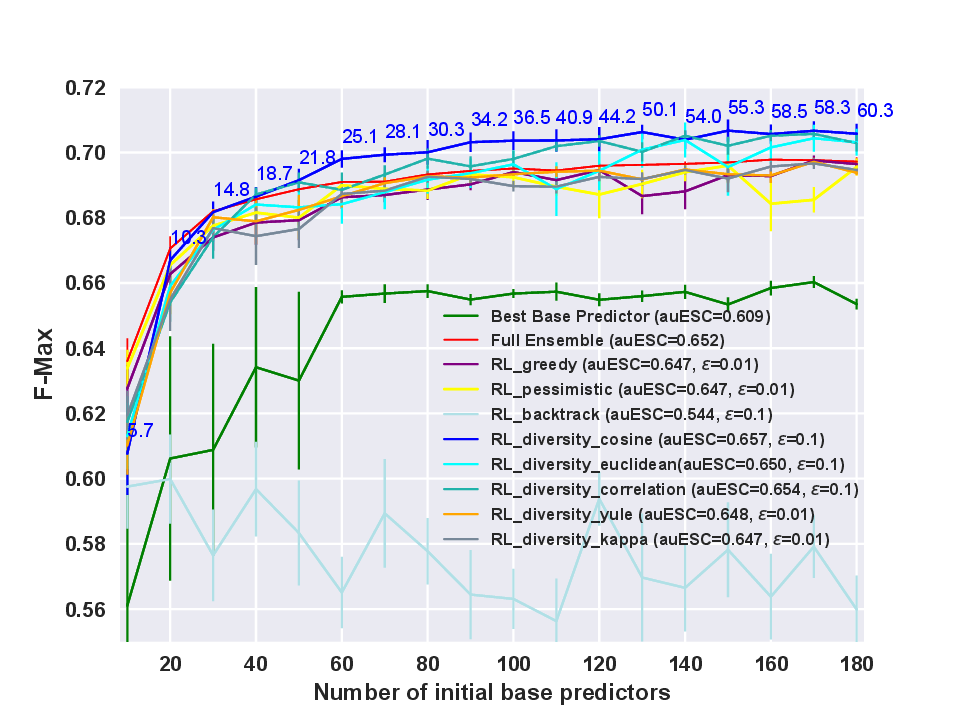}
\caption{Performance of ensembles produced by several algorithms, including the diversity-incorporating RL-based ones proposed here, on the \emph{C. remanei} SS data set as the number of base predictors is increased. Also shown for reference are the performances of the full ensemble comprising of all the base predictors, as well as the best one among them. The best performing method out of \emph{diversity1} and \emph{diversity2}, and value of $\epsilon$ (shown within parentheses), was used to generate the performance curve for each diversity measure. Standard errors are calculated over ten repetitions of each respective experiment. The numbers on top of the RL\_diversity\_cosine (blue) curve represent the average sizes of the ensembles selected by the algorithm at various numbers of initial base predictors.}
\label{RL_comparison_ramenei}
\end{figure*}

\subsection{Comparison of RL-based ensemble selection algorithms}
In previous work \cite{Stanescu:2017}, we proposed several RL-based ensemble selection algorithms, which performed comparably to the full ensemble (FE)
consisting of all the base predictors, especially considering that the ensembles produced by these algorithms were much smaller than FE, i.e. more parsimonious. Thus, based on the observations made in the previous subsection, we compared the performance of these ensembles and baselines with that of our diversity-incorporating RL ensembles on the various test data sets. 

Figure \ref{RL_comparison} provides an overview of the results of this comparison. It can be seen that almost all the ensembles, whether FE- or RL-based ones, performed substantially better than the best base predictor (light green bars), an essential requirement for ensemble performance. The only exception to this was the RL\_backtrack algorithm, which is prone to overfitting due to its relatively complex lattice search algorithm. Furthermore, Figure \ref{RL_comparison} shows that none of our previous RL algorithms outperformed the full ensemble in terms of the absolute auESC score.

Next, we assessed how the performance of the diversity-incorporating RL ensembles compared with that of the full ensemble. For the PF data sets (Figure \ref{RL_comparison}(a)), the RL\_diversity ensembles perform at a level comparable to our previously proposed RL approaches, although slightly lower than that of the full ensemble. On the other hand, several of the RL\_diversity ensembles perform better than FE (red bars) on SS data sets, and transitively better than our previous RL approaches. For instance, RL\_diversity\_cosine (blue bars) outperforms FE for all the PF data sets except \emph{A. thaliana}, which is likely due to the unbalanced nature of our data sets (many more negatives than positives). In situations like these, similarity measures like cosine are generally  more meaningful. RL\_diversity\_correlation (green bars) and RL\_diversity\_euclidean (light blue bars) produce such a performance for a couple of data sets each. This indicates that these variations of the diversity-incorporating algorithms are able to utilize the information in these much larger than PF data sets to derive more accurate ensembles. 

{\small
\begin{table*}[t!]
\begin{tabular}{l|l|ccc|ccc}
RL algorithm	& auESC    & size\_ratio@60 & size\_ratio@120 & size\_ratio@180 & perf\_ratio@60 & perf\_ratio@120 & perf\_ratio@180 \\ \hline
RL\_greedy                 & 0.647 & 0.761 & 0.676 & 0.618 & 0.993 & 0.998 & 0.999 \\
RL\_pessimistic            & 0.647 & 0.497 & 0.292 & 0.195 & 0.999 & 0.987 & 0.998 \\
RL\_backtrack              & 0.545 & 0.115 & 0.069 & 0.036 & 0.818 & 0.853 & 0.803 \\\hline
{\bf RL\_diversity\_cosine}      & {\bf 0.657} & {\bf 0.418} & {\bf 0.368} & {\bf 0.335} & {\bf 1.010} & {\bf 1.012} & {\bf 1.012} \\
RL\_diversity\_euclidean   & 0.650 & 0.389 & 0.358 & 0.326 & 0.990 & 0.998 & 1.008 \\
RL\_diversity\_correlation & 0.654 & 0.456 & 0.357 & 0.316 & 0.996 & 1.011 & 1.008 \\
RL\_diversity\_yule        & 0.648 & 0.772 & 0.642 & 0.647 & 0.994 & 0.998 & 0.995 \\
RL\_diversity\_kappa       & 0.647 & 0.995 & 0.995 & 0.996 & 0.995 & 0.995 & 0.996      
\end{tabular}
\caption{Statistics for the performance curves shown in Figure \ref{RL_comparison_ramenei}.
The ratios of the sizes and performances of the ensembles produced by the various RL algorithms to those of the full ensemble (FE) are shown at representative initial base predictor set sizes of 60, 120, and 180.}
\label{parsimony_table_ramenei}
\end{table*}
}

In contrast, the RL\_diversity variations based on supervised diversity measures didn't perform better than FE on any of the data sets, either PF or SS. Thus, even though in aggregate, the two types of measures perform statistically similarly (Section 4.1), there are individual differences. This indicates that while established measures like $Q$ and $\kappa$ are useful for studying ensembles, their direct use for learning ensembles using algorithms like ours may not necessarily be effective. In such cases, it may be more meaningful to use simpler unsupervised measures that are easier to compute and less likely to overfit.

\subsection{Relationship between ensemble performance and parsimony}
As was pointed out earlier,  an advantage of ensemble selection is that it can not only help learn accurate ensembles, but the relatively smaller size (\emph{parsimony}) of these ensembles can aid interpretation or reverse engineering efforts. Our previous work \cite{Stanescu:2017} illustrated that RL can indeed help achieve this parsimony goal while not compromising too much on performance. We assessed if and to what extent the diversity-incorporating ensembles can help in this direction. However, since presenting the results of this assessment for all our test data sets will take too much space, we illustrate the results using the \emph{C. remanei} SS data set, which produces the overall highest performance across all ensemble algorithms (Figure \ref{RL_comparison}(b)).

Figure \ref{RL_comparison_ramenei} shows the comparative performance of ensembles produced by several algorithms, including the diversity-incorporating RL-based ones proposed here, on this data set. Consistent with the results for this data set shown in Figure \ref{RL_comparison}(b), several RL\_diversity algorithms do perform better than the full ensemble (red curve), as shown by higher performance curves. In particular, RL\_diversity with the cosine (blue curve) and correlation (green curve) diversity measures (almost) consistently outperform the full ensemble, especially after about $40$ initial base predictors. Thus, in addition to at the aggregate (auESC) level (Figure \ref{RL_comparison}(b)), this result shows that diversity-incorporating ensembles can also work well at a data set level and for a range of numbers of initial base predictors.

However, as discussed above, performance is only a part of the story. The other part is how much parsimony the RL ensembles are able to achieve relative to the full ensemble. The numbers on top of the RL\_diversity\_cosine (blue) curve in Figure \ref{RL_comparison_ramenei} represent the average sizes of the ensembles selected by this algorithm with various numbers of initial base predictors. It can be seen that RL\_diversity\_cosine not only performs better than the full ensemble, but also does this consistently with much smaller ensembles, often nearly only $30-40\%$ of the size of the full ensemble. 

This can also be seen more generally in Table \ref{parsimony_table_ramenei}, which reports the ratios of the sizes and performances of the ensembles produced by the various RL algorithms to those of the full ensemble (FE) at representative initial base predictor set sizes of 60, 120, and 180. It can be seen here also that the RL\_diversity\_cosine ensembles consistently performs better than the full ensemble and ensembles from other RL-based approaches while maintaining small ensemble sizes. Other unsupervised diversity measures (euclidean and correlation) also do well in this direction, but supervised ones (yule and kappa) don't, once again confirming the observation made above.

Overall, the above results demonstrate that diversity-incorporating RL ensemble selection can help infer accurate as well as parsimonious ensembles. This can be attributed to the approach's ability to systematically explore diverse base predictors whose addition to the ensemble can aid ensemble performance and parsimony.

\section{Conclusions}
In this paper, we describe a set of algorithms to incorporate ensemble diversity into our reinforcement learning (RL) framework for ensemble selection. These algorithms were implemented using a variety of diversity incorporation methods and ensemble diversity measures, and were rigorously evaluated on several challenging problems and associated data sets. Several of these implementations produced more accurate ensembles than those that don't explicitly consider diversity, such as ones combining all the base predictors and those based on our previous RL approaches, especially for larger data sets. Perhaps even more importantly, these diversity-incorporating  ensembles were much smaller in size, i.e., more parsimonious, than the latter types of ensembles. This can eventually aid the interpretation or reverse engineering of predictive models assimilated into the resultant ensemble(s).

Our proposed approaches can be improved upon by considering other methods of incorporating diversity, as well as measures for quantifying it. Particularly, the lack of success of specialized (supervised) ensemble diversity measures within our framework needs to be investigated. Also, it will be beneficial to objectively assess if and how the parsimonious nature of our ensembles can actually aid interpretation, as we hypothesize. Finally, as with several other uses of reinforcement learning \cite{Sutton:1998}, computational efficiency is a challenge for us as well. Thus, there is a need for more efficient RL algorithms and implementations, such as those based on parallelization and high-performance computing, that can enhance the efficiency of our approaches as well. 

\begin{acks}
This work was partially supported by NIH grant \# R01-GM114434 and an IBM faculty award to GP. It was also supported by the computational resources and staff expertise provided by Scientific Computing at the Icahn School of Medicine at Mount Sinai. 
\end{acks}

\bibliographystyle{ACM-Reference-Format}
%\bibliography{bibliography_ana,icdm2013}
\bibliography{kdd_2018_ana.bbl}
\end{document}